\documentclass{article}
\usepackage[letterpaper, margin=1in]{geometry}

\title{TASRA: a Taxonomy and Analysis of \\
Societal-Scale Risks from AI \\ }
\usepackage{natbib}
\usepackage[utf8]{inputenc} 
\usepackage[T1]{fontenc}    

\usepackage{hyperref}       
\PassOptionsToPackage{hyphens}{url}\usepackage{hyperref}
\usepackage{hyperref}
\hypersetup{pdftex,colorlinks=true,allcolors=blue}
\usepackage{hypcap}
\usepackage{url}            

\usepackage{booktabs}       
\usepackage{amsfonts}       
\usepackage{nicefrac}       
\usepackage{microtype}      
\usepackage{amsmath}
\usepackage{amsthm}
\usepackage{amssymb}
\usepackage{float}
\usepackage{graphicx}
\usepackage{xcolor}
\usepackage{amsmath}
\usepackage{amsthm}
\usepackage{amssymb}
\usepackage{wrapfig}
\usepackage[capitalise]{cleveref}

\makeatletter
\newcommand{\vast}{\bBigg@{3}}
\newcommand{\Vast}{\bBigg@{4}}
\makeatother

\makeatletter
\newcommand{\bidef}[3]{
  \protected@write\@auxout{}{%
    \global\string\@namedef{#1:#2}{#3}%
  }%
}

\newcommand{\biref}[3]{
  \@ifundefined{#1:#2}{??}{\@nameuse{#1:#2}}%
}
\makeatother

\usepackage{todonotes}
\usepackage{soul}
\usepackage{marginnote} %

\newif\ifcomments 

\newif\iffootnote
\let\Footnote\footnote
\renewcommand\footnote[1]{\begingroup\footnotetrue\Footnote{#1}\endgroup}

\newcommand\setupcomments[4]{
\expandafter\def\csname#1says\endcsname##1{\ifcomments{\color{#3}[#2: ##1]}\fi}
\expandafter\def\csname#1predicts\endcsname##1{\ifcomments{\color{#3}[#2 predicts: ##1 \#}\fi}
\expandafter\def\csname#1color\endcsname##1{\ifcomments{\color{#3}##1}\fi}
\expandafter\def\csname#1adds\endcsname##1{\ifcomments{\color{#3}##1}\fi}
\expandafter\def\csname#1rems\endcsname##1{\ifcomments{\color{#3}\st{##1}}\fi}
  \expandafter\def\csname#1box\endcsname##1{\ifcomments\vspace{1ex}\todo[inline,#4]{#2 says: ##1}{}\fi}

  \expandafter\def\csname#1margin\endcsname##1{\ifcomments\marginnote{\todo[inline,#4]{#2 says: ##1}{}}\fi}
  \expandafter\def\csname#1section\endcsname##1{\ifcomments\todo[#4]{#2 says: ##1}{}\fi}
  \expandafter\def\csname#1todo\endcsname##1{\ifcomments\marginnote{\todo[inline]{#2 says: TODO: ##1}{}}\fi}

  \expandafter\def\csname#1demo\endcsname{\ifcomments
\noindent
$\backslash$#1says $\to$ \csname#1says\endcsname{example inline comment from #2.}\\
$\backslash$#1adds $\to$ \csname#1adds\endcsname{example addition by #2.}\\
$\backslash$#1rems $\to$ \csname#1rems\endcsname{example removal by #2.}\\
$\backslash$#1box $\to$ 
    \csname#1box\endcsname{example box comment from #2}
\fi}

}

\commentstrue
\usepackage[linesnumbered,ruled]{algorithm2e}
\usepackage{framed}
\usepackage{quoting}
\colorlet{shadecolor}{blue!10}
\newenvironment{shadedquotation}
 {\begin{shaded*}
  \quoting[leftmargin=0pt, vskip=0pt]
 }
 {\endquoting
 \end{shaded*}
}

\usepackage{comment}
\usepackage{float}

\usepackage{scrextend} 
\usepackage{environ} 

\NewEnviron{story}[3]{
\vspace{1ex}
\label{story:#1}\bidef{story}{#1}{#2}
\begin{shadedquotation}
\textbf{#3}
\BODY
\end{shadedquotation}
\vspace{1ex}
}
\newcommand{\storyref}[1]{\hyperref[story:#1]{\biref{story}{#1}}}

\commentstrue
\setupcomments{ac}{AC}{purple!100!black}{color=blue!30}
\setupcomments{sr}{SR}{green!70!black}{color=green!10}

\usepackage{array}
\newcolumntype{L}[1]{>{\raggedright\let\newline\\\arraybackslash\hspace{0pt}}p{#1}}
\newcolumntype{C}[1]{>{\centering\let\newline\\\arraybackslash\hspace{0pt}}p{#1}}
\newcolumntype{R}[1]{>{\raggedleft\let\newline\\\arraybackslash\hspace{0pt}}p{#1}}

\author{
  Andrew Critch\thanks{Center for Human-Compatible Artificial Intelligence, Department of Electrical Engineering and Computer Sciences, UC Berkeley} \\
  \texttt{critch@eecs.berkeley.edu} \\
   \and
     Stuart Russell$^*$ \\
  \texttt{russell@cs.berkeley.edu} \\
}

\begin{document}

\maketitle

\begin{abstract}
While several recent works have identified societal-scale and
extinction-level risks to humanity arising from artificial intelligence, few have attempted an {\em exhaustive taxonomy} of such risks.  Many exhaustive taxonomies are possible, and some are useful---particularly if they reveal new risks or practical approaches to safety. This paper explores a taxonomy based on accountability: whose actions lead to the risk, are the actors unified, and are they
deliberate?  We also provide stories to illustrate how the various risk types could each play out, including risks arising from unanticipated interactions of many AI systems, as well as risks from deliberate misuse, for which combined technical and policy solutions are indicated.  

\end{abstract}

\section{Introduction} 
A few weeks ago, a public statement was signed by leading scientists and executives in AI, stating that ``Mitigating the risk of extinction from AI should be a global priority alongside other societal-scale risks such as pandemics and nuclear war'' \citep{cais2023statement}.  
This represents a significant increase in coordinated concern for human extinction risk arising from AI technology, and implies more generally that catastrophic societal-scale risks from AI should be taken as a serious concern.  In consonance, just a few days ago US President Joe Biden and UK Prime Minister Rishi Sunak expressed an agreement to ``work together on AI safety, including multilaterally'', citing that ``last week, the pioneers of artificial intelligence warned us about the scale of the challenge'' \citep{bidensunak2023press}.

Meanwhile, in recent years national governments throughout the world have begun to address societal-scale risks from AI.  In 2018, Chinese leader Xi Jinping exhorted the attendees of the World AI Conference to ``make sure that
artificial intelligence is safe, reliable and controllable''.  Since then, several AI governance initiatives have emerged in China \citep{sheehan2021china}, including specific measures for generative AI services drafted in April of this year \citep{china2023measures, huang2023translation}.  In Europe, the proposed European Union AI Act began in large part as a 
response to concerns that AI systems may pose risks to the
safety and fundamental rights of humans \citep{eu2021ai}. 
In the US, last year the White House issued a Blueprint for an AI Bill of Rights \citep{whitehouse2022blueprint}, addressing ``challenges posed to democracy today'' by ``the use of technology, data, and automated systems in ways that threaten the rights of the American public.''



\begin{figure}[H]
\centering
  \caption{\label{figure:nist} Purple and orange annotations on Figure 2 of the NIST ``AI Risk Management Framework: Initial Draft'', indicating what we consider to be ``societal-scale risks''.}
  \includegraphics[width=0.9\textwidth]{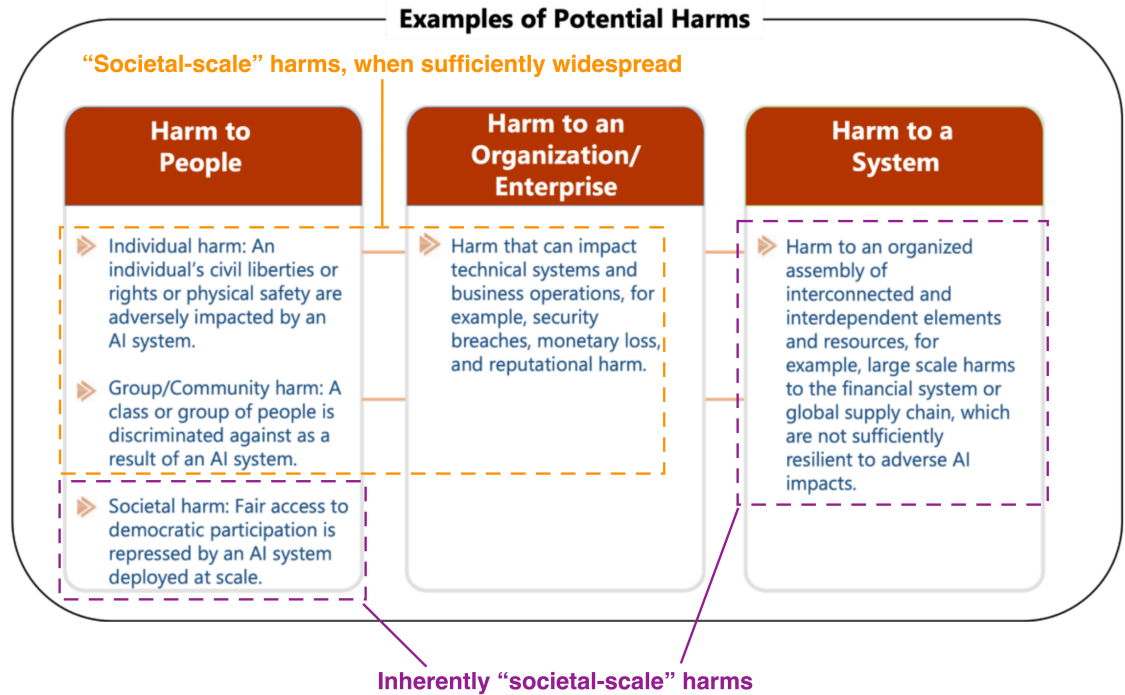}
  \label{NIST-figure}
\end{figure}

Harms occurring at the scale of individual persons may be distinguished from harms occurring on the scale of an entire society, which we call \emph{societal-scale harms}.  This distinction can also be seen somewhat in last year's report
from the US National Institute of Standards and Technology proposing an ``AI risk management framework'' \citep{nist2022airmf}, which
distinguished individual harms from ``societal harm'' and
``harms to a system [...], for example, large scale harms to the
financial system or global supply chain''; see Figure \ref{NIST-figure}.  Harms to individuals and groups should also be considered ``societal-scale'' when sufficiently widespread.

\vspace{-6ex}

How should societal-scale risks be addressed in technical terms?  So far, most research papers addressing societal-scale and existential risks have focused on misalignment of a single advanced AI system.  In a recent blog post, \citet{bengio2023rogue} lays out a clear and concise logical argument for this case, entitled ``How Rogue AIs may Arise''.  However, while misalignment of individual systems remains a problem, it is not the only source of societal-scale risks from AI, and extinction risk is no exception.  Problems of racism, misinformation, election interference, and other forms of injustice are all risk factors affecting humanity's ability to function and survive as a healthy civilization, and can all arise from interactions between multiple systems or misuse of otherwise ``aligned'' systems.  And, while \citet{russell2019human} has offered the single-human/single-machine framing as a ``model
for the relationship between the human race and its machines, each
construed monolithically,'' this monolithic view of AI technology is not enough: safety requires analysis of risks at many scales of organization simultaneously.  Meanwhile, \citet{ng2023extinction} together have called for a better articulation of concrete risks from AI, including extinction risk.  

In this paper, we expand our focus somewhat from the implicit assumption that societal-scale harms must result from a single misaligned system, and begin to analyze societal-scale risks in accordance with the decision tree in Figure 2 below:

\begin{figure}[H]
\centering
  \caption{\label{figure:decision} An exhaustive decision tree for classifying societal-scale harms from AI technology}
  \vspace{1ex}
  \includegraphics[width=0.9\textwidth]{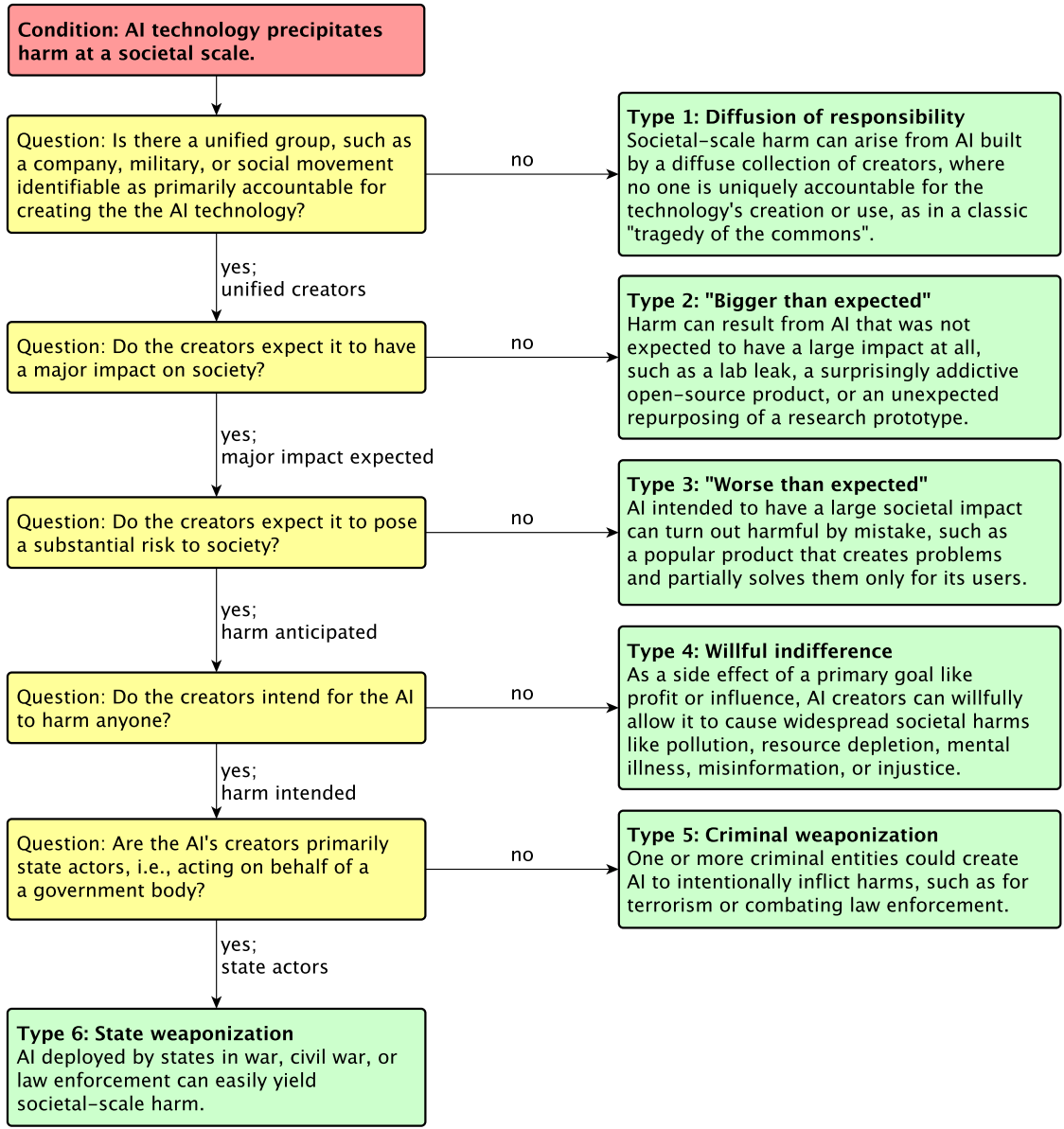}
\end{figure}

Safety engineers often carry out a \emph{fault tree
analysis}~\citep{watson1961launch, mearns1965fault,  lee1985fault} as a way to ensure they have covered all
possible failures.  The root of a fault tree is the condition to be
avoided and
each branch tests some condition.  As long as the branches from each
node are logically exhaustive, the leaves necessarily cover all
possible circumstances. Typically branches test whether a given subsystem is working correctly or not, but can also
test more general conditions such as the ambient temperature or whether the system is undergoing testing.  The decision tree in Figure~\ref{figure:decision} above follows the same basic principle to produce an exhaustive taxonomy.

Exhaustiveness of a taxonomy is of course no guarantee of usefulness. For example, an analysis based on whether the day of the month is a prime number would yield an exhaustive two-leaf taxonomy while providing zero analytical benefit. A taxonomy
is only useful to the extent that it reveals new risks or recommends helpful interventions.

To that end, we have chosen an exhaustive taxonomy based on accountability: whose
actions led to the risk,  were they unified, and were they deliberate? Such a taxonomy may
be helpful because it is closely tied to the important questions of
where to look for emerging risks and what kinds of policy
interventions might be effective.  This taxonomy in particular surfaces risks arising from unanticipated interactions of many AI systems,
as well as risks from deliberate misuse, for which combined technical
and policy solutions are needed.  

Many other taxonomies are possible and should be explored.  A previous taxonomy of \citet{yampolskiy2015taxonomy} also examined sources of AI risk arising intentionally, by mistake, or from a system's environment, either pre-deployment or post-deployment.  While useful, Yampolskiy's taxonomy was non-exhaustive, because it presumed a unified intention amongst the creators of a particular AI system.  In reality, no well-defined ``creator's intent'' might exist if multiple AI systems are involved and built with different objectves in mind.

\subsection{Related work and historical context}

Historically, the risk posed to humanity by advanced AI systems was first recognized in fiction, by authors such as
Samuel \citet{butler1863darwin} and Karel \citet{capek1921rur}.  Later, warnings were also expressed by computer scientists such as Alan \citet*{turing1951can, turing1951intelligent} and Norbert
\citet{wiener1960some}, with Wiener pinning risk on the difficulty of ensuring
``the purpose put into the machine'' would be  aligned with
actual human preferences, and I.~J.~\citet{good1966speculations} highlighted the
additional threat of rapid, recursive self-improvement leading to a
loss of control.  

In this century, many have examined existential
risk from superintelligent
machines~\citep{hibbard2001super,yudkowsky2008artificial,barrat2013artificial,bostrom2014superintelligence, yampolskiy2015taxonomy}
and various technical approaches have been suggested to address it,
particularly in the area of
AI alignment~\citep{soares2014aligning,russell2014white,hadfield2016cooperative,amodei2016concrete,russell2019human}.   

\section{Types of Risk}
Here we begin our analysis of risks organized into six risk types, which constitute an exhaustive decision tree for classifying societal harms from AI or algorithms more broadly.  Types 2-6 will classify risks with reference to the intentions of the AI technology’s creators, and whether those intentions are being well served by the technology. Type~1, by contrast, is \emph{premised} on no single institution being primarily responsible for creating the problematic technology.  Thus, Type~1 serves as a hedge against the taxonomy of Types 2-6 being non-exhaustive.

\subsection{Type~1: Diffusion of responsibility}

Automated processes can cause societal harm even when no one in particular is primarily responsible for the creation or deployment of those processes \citep{zwetsloot2019thinking}, and perhaps even as a result of the absence of responsibility. The infamous ``flash crash'' of 2010 is an instance of this: numerous stock trading algorithms from a variety of companies interacted in a fashion that rapidly devalued the US stock market by over 1 trillion dollars in a matter of minutes.  Fortunately, humans were able to intervene afterward and reverse the damage, but that might not always be possible as AI technology becomes more powerful and pervasive.  

Consider the following fictional story, where the impact of unemployment on crime rates \citep{raphael2001identifying} is exacerbated by a cycle of algorithmic predictions:


\begin{story}{pessimism}{Self-Fulfilling Pessimism}{Story 1a: Self-Fulfilling Pessimism.}  Scientists develop an algorithm for predicting the answers to questions about a person, as a function of freely available and purchasable information about the person (social media, resumes, browsing history, purchasing history, etc.).  The algorithm is made freely available to the public, and employers begin using the algorithm to screen out potential hires by asking, ``Is this person likely to be arrested in the next year?''  Courts and regulatory bodies attempt to ban the technology by evoking privacy norms, but struggle to establish cases against the use of publicly available information, so the technology broadly remains in use.

Innocent people who share certain characteristics with past convicted criminals end up struggling to get jobs, become disproportionately unemployed, and correspondingly more often commit theft to fulfill basic needs.  Meanwhile, police also use the algorithm to prioritize their investigations, and since unemployment is a predictor of property crime, the algorithm leads them to suspect and arrest more unemployed people.  Some of the arrests are talked about on social media, so the algorithm learns that the arrested individuals are likely to be arrested again, making it even more difficult for them to get jobs.  A cycle of deeply unfair socioeconomic discrimination begins.
\end{story}

In the story above, a subset of humanity becomes unfairly disempowered, both economically and legally.  It is possible, we claim, for \textit{all of humanity} to become similarly disempowered.  How?

Consider that many systems of production and consumption on Earth currently operate entirely without human involvement, while producing side effects for humans and other life.  For instance, algal blooms consume energy from the sun and materials from the surrounding ocean, and as a side effect they sometimes produce toxins that are harmful to other sea life as well as human swimmers.  It is important to consider the possibility that artificially intelligent systems, in the future, could also sustain fully self-contained loops of production and consumption that would yield negative side effects for humanity.  The following diagram illustrates how a few industries, if fully automated through AI technology, could operate in a closed loop of production (and consumption) without any other inputs:

\begin{figure}[H]\label{figure:production}
  \caption{A hypothetical self-contained ``production web'' of companies operating with no human involvement; such a production web would make it possible to completely decouple economic activities from serving human values.}
  \vspace{1ex}
  \includegraphics[width=\textwidth]{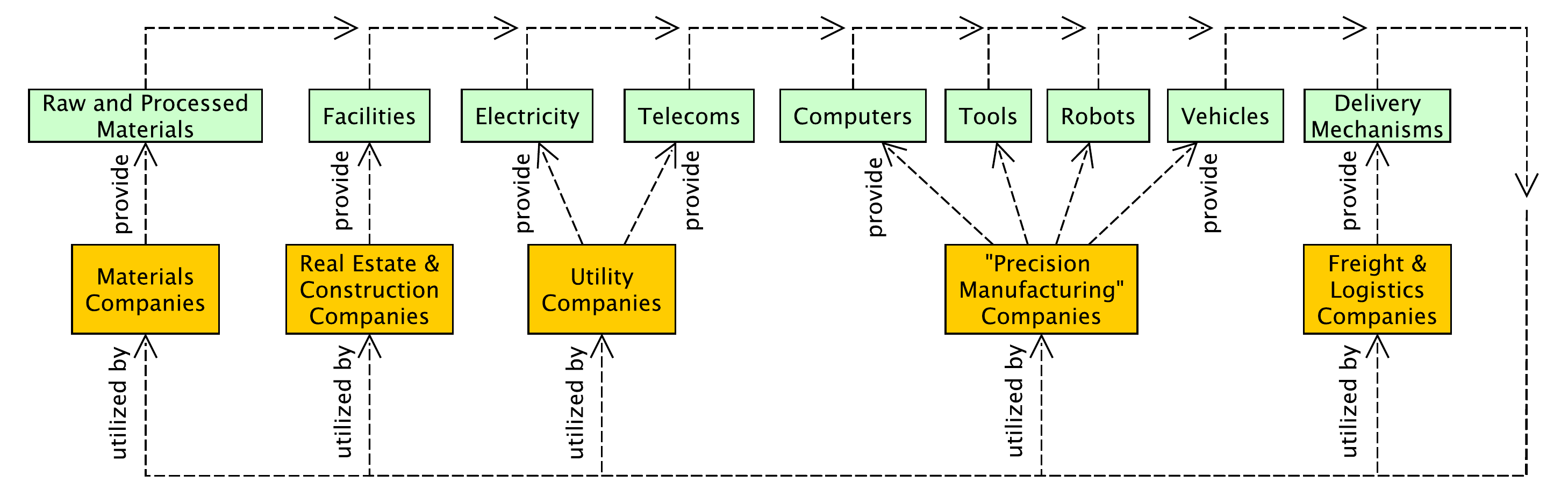}
\end{figure}


Could such a self-contained ``production web'' ever pose a threat to humans?  One might argue that, because AI technology will be created by humanity, it will always serve our best interests.  However, consider how many human colonies have started out dependent upon a home nation, and eventually gained sufficient independence from the home nation to revolt against it.  Could humanity create an ``AI industry'' that becomes sufficiently independent of us to pose a global threat?

It might seem strange to consider something as abstract or diffuse as an \emph{industry} posing a threat to the world.  However, consider how the fossil fuel industry was built by humans, yet is presently very difficult to shut down or even regulate, due to patterns of regulatory interference exhibited by oil companies in many jurisdictions \citep{carpenter2013preventing, dal2006regulatory}.  The same could be said for the tobacco industry for many years \citep{gilmore2019tobacco}.  The ``AI industry'', if unchecked, could behave similarly, but potentially much more quickly than the oil industry, in cases where AI is able to think and act much more quickly than humans.

Finally, consider how species of ants who feed on acacia trees eventually lose the ability to digest other foods, ending up ``enslaved'' to protecting the health of the acacia trees as their only food source \citep{yong2013trees}.  If humanity comes to depend critically on AI technology to survive, it may not be so easy to do away with even if it begins to harm us, individually or collectively.

For an illustration of how this might happen, consider the story below:

\begin{story}{production}{Production Web}{Story 1b: The Production Web.} Someday, AI researchers develop and publish an exciting new algorithm for combining natural language processing and planning capabilities. Various competing tech companies develop ``management assistant'' software tools based on the algorithm, which can analyze a company's cash flows, workflows, and communications to recommend more profitable business decisions that also yield positive PR and networking opportunities for managers.  It turns out that managers are able to automate their own jobs almost entirely, by having the software manage their staff directly.  Software tools based on variants of the algorithm sweep through companies in nearly every industry, automating and replacing jobs at various levels of management, sometimes even CEOs.  One company develops an ``engineer-assistant'' version of the assistant software, capable of software engineering tasks, including upgrades to the management assistant software.  

Within a few years, it becomes technologically feasible for almost almost any human job to be performed by a combination of software and robotic workers that can operate more quickly and cheaply than humans, and the global job market gradually begins to avail of this possibility.  A huge increase in global economic productivity ensues. 
Despite the massive turnover in the job market, average quality of life also improves in almost every country, as products and services become cheaper to produce and provide.  Most job losses come with generous severance packages, sometimes enough for a full retirement.  

Companies closer to becoming fully automated achieve faster turnaround times, deal bandwidth, and creativity of business-to-business negotiations.  Some companies idealistically cling to the idea the human workers must remain integral to their operations, however, they quickly fall behind because they simply can't provide products and services as cheaply as their fully automated competitors.  Eventually, almost all companies either fail and shut down or become fully automated.

An interesting pattern of trade begins to emerge between a conglomerate of automated companies in the materials, real estate, construction, utilities, and freight and logistics industries, along with a new generation of ``precision manufacturing'' companies that can use robots to build almost anything if given the right materials, a place to build, some 3d printers to get started with, and electricity. Together, these companies sustain an increasingly self-contained and interconnected ``production web'' that can operate with no input from companies outside the web, while providing an impressive swath of cheap products and services to the rest of the world.
 
The objective of each company in the production web could loosely be described as an amorphous combination of profitability, size, market share, and social status, all learned by emulating the decision-making of human business leaders.  These objectives are implemented as large and opaque networks of parameters that were tuned and trained to optimize the inferred objectives of human business leaders during the early days of the management assistant software boom.

At this point, the story hits an inflection point that is difficult for the characters in it to perceive.  In short, the world begins to change in a way that renders the production web a harmful rather than helpful presence, but the change happens so gradually that collective action against it is difficult to precipitate, and eventually the change is irreversible.  The details of this change --- or rather, just one way it could play out --- constitute the remainder of the story...

First, human leaders in more conservative jurisdictions struggle to keep track of how the production web companies are producing so many products and so cheaply, and without easily accessible human-legible paper trails, auditing attempts glean little insight.

As time progresses, it becomes increasingly unclear---even to the concerned and overwhelmed Board members of the fully mechanized companies of the production web---whether these companies are serving or merely appeasing humanity.  We eventually realize with collective certainty that the companies have been trading and optimizing according to objectives misaligned with preserving our long-term well-being and existence, but by then their facilities are so pervasive, secure, and necessary for serving our basic needs that we are unable to stop them from operating.  With no further need for the companies to appease humans in pursuing their production objectives, less and less of their activities end up benefiting humanity.  
Eventually, human-critical resources (e.g., arable land, drinking water, atmospheric oxygen) are depleted and climate conditions are compromised at an alarming rate, threatening humanity’s very existence.  

In the end, humanity is faced with a difficult collective action problem: deciding when and in what way to physically stop the production web from operating.  In the best case, a shutdown is orchestrated, leading to decades of economic dislocation, deprivation, and possibly famine.  In the worst case, military-level conflict emerges between humanity and the fully automated companies, or humanity simply perishes before mounting a coordinated defense.
\end{story}

\paragraph{Analysis.} In the story above, rapidly operating institutions tended toward trade and interaction with other rapidly operating institutions, thus yielding a collective tendency towards ``closing the loop'' on production and consumption by fully automated companies.  Abstractly, it may be summarized as follows:
\begin{enumerate}
    \item AI technology proliferated during a period when it was beneficial and helpful to its users.
    
    \item There was a gradual handing-over of control from humans to AI systems, driven by competitive pressures for institutions to (a) operate more quickly through internal automation, and (b) complete trades and other deals more quickly by preferentially engaging with other fully automated companies.
    
    \item Humans were not able to collectively agree upon when and how much to slow down or shut down the pattern of technological advancement.
    
    \item Once a closed-loop ``production web'' had formed from the competitive pressures in 2(a) and 2(b), the companies in the production web had no production- or consumption-driven incentive to protect human well-being, and eventually became harmful.
    
\end{enumerate}
 
\paragraph{What can be done?}  What kinds of checks and balances are needed to keep stories like these entirely in the domain of science fiction?  For one thing, when the activities of multiple agents are collectively giving rise to risks or harms, new behavior norms for the agents are needed to steer them collectively away from the harmful pattern.  Indeed, both stories above included shortfalls in regulatory efforts.  To prevent such scenarios, effective regulatory foresight and coordination is key.

Agriculture provides an interesting precedent for regulation.  Historically, agricultural products---like algorithms---have had the potential to be replicated and misused, leading to societal harm, including harms not easily perceptible by any one individual.  For instance, we now know that small amounts of lead in food can yield a slow accumulation of mental health problems, even when the amount of lead in any particular meal is imperceptible to an individual consumer.  Widespread degradation of mental health can lead to many other large-scale harms, including the breakdown of institutions that depend on mentally healthy constituents to function.  Today, this problem is avoided through regulation.  The United States Food and Drug Administration (FDA) relies on the rigorous classification and testing of food and drugs to protect the public from health risks, as well as on stringent requirements for grocery stores and pharmacies to maintain legible records.  

By contrast, there is currently no such pervasive and influential regulatory body for algorithms in the United States.  At present, tech companies follow their own internal policies for protecting users, with little external oversight of interactive algorithms and their effects on people.  However, there has been some discussion in the Senate regarding the creation of a federal agency for this purpose \citep{ussenate2023hearing, ussenate2023oversight}. 
Also, the National Institute of Standards and Technology (NIST) is presently compiling non-regulatory standards for AI technology, but these are suggestions rather than legally enforced requirements.  The General Data Protection Regulation (GDPR) is enforceable in the EU, and California's Consumer Privacy Act (CCPA) applies to protect consumers in California; however, unless such regulations are adopted more widely, they may simply serve to determine \emph{where} harmful algorithms operate, rather then \emph{whether}.
 
How would we even begin to classify and test \emph{algorithms} for regulatory oversight, the way foods are classified and tested?  Many approaches could make sense here.  One that stands out is a language called UML (Unified Modeling Language) specifically designed for documenting and diagramming the architecture of IT systems at an abstract level, including workflow interactions with humans. Perhaps a UML-like language could be used to establish standards for classifying and regulating algorithms.  

Thinking purely quantitatively, oversight could also be triggered entirely on the basis of large computational resource expenditures.  Companies could be required to produce auditable reports of how they use computing and communications resources, just as they are already required to report on their usage of money or controlled substances \citep{jackson2022compute}.  In testimony to the U.S. Senate Committee on Armed Services in April of this year, RAND CEO Jason Matheny recently advocated for ``Defense Production Act authorities to require companies to report the development or distribution of large AI computing clusters, training runs, and trained models (e.g. >1,000 AI chips, >1027 bit operations, and >100 billion parameters, respectively)'' \citep{matheny2023testimony}.  Auditing is discussed more under risk Type~4 below, specifically with regards to the interpretability of AI algorithms.
 
More than individual company audits will be necessary to prevent large-scale interactions between diffuse collections of companies from leading to negative externalities for society.  Humanity, collectively, will eventually need to enable one or more regulatory institutions to view the interaction of computing and communications systems at a global scale, to detect if and when those global interactions are beginning to lead the world down a harmful path that no individual company might be responsible for preventing (like in the stories), or even capable of noticing.

Who or what oversight bodies should be privy to such a ``global report of computing and communications activity''?  It might be the entire public, one or more government agencies, an international NGO, or a professional standards organization.  In all cases, help will be needed from domain experts to assess potential risks that could arise from the worldwide aggregate behavior of algorithms.  A new discipline that essentially unifies control theory, operations research, economics, law, and political theory will likely be needed to make value judgements at a global scale, irrespective of whether those judgements are made by a centralized or distributed agency.
 
Where could we begin to develop such a unified discipline?  At a technical level, one might start by developing simplified mathematical models of the \emph{sociotechnical context} in which AI algorithms operate, perhaps leaning on UML or another systems-level modelling language for inspiration.  A few micro-scale examples of this are given under Type~4; as a macro-scale example, perhaps a UML-like model of the global economy in the story above might yield Figure \ref{figure:production} as a sub-diagram of a larger production diagram.  

In summary, the following three problems need to be addressed:
\begin{itemize}
\item \textbf{Regulatory problem:} Algorithms and their interaction with humans will eventually need to be regulated in the same way that food and drugs are currently regulated.

\item \textbf{Oversight problem:} One or more institutions will be needed to oversee the worldwide behavior and impact of non-human algorithms, hereby dubbed ``the algorithmic economy.''  This should include assessments of whether humanity retains the ability to shut down or redirect the algorithmic economy, with an eye toward the risk of self-contained production webs developing over time.

\item \textbf{Technical problem:} A new technical discipline will be needed to classify and analyze the sociotechnical context of algorithms for the purposes of oversight and regulation.
\end{itemize}
\subsection{Type~2: ``Bigger than expected'' AI impacts}

The \storyref{pessimism} and \storyref{production} stories above already illustrate how the scope of actions available to an AI technology can be greatly expanded when the technology is copied many times over, or modified relative to the likely intentions of its initial creators. However, impact on an unexpectedly large scale can occur even if only one team is responsible for creating the technology.  

The following story illustrates how a new AI technology can yield a negative societal-scale impact as a result of its developers failing to adequately understand the mechanism by which its societal-scale impact would occur:
 
\begin{story}{hatespeech}{Hate Speech Leak}{Story 2a: Hate Speech Leak.} A social media company decides to develop a content moderation tool for flagging instances of hate speech. For testing purposes, AI researchers train a natural language text generator to produce a large volume of artificial hate speech, which turns out to be quite creative in the hateful arguments it generates, and helps the company to develop very robust hate-speech detection and flagging algorithms.  But one day the hate speech corpus is accidentally leaked onto the Internet, yielding a highly negative global impact where persons looking to incite hatred begin re-using its statements as ``scientifically proven insults.''\footnote{The Microsoft Tay fiasco shows how much trouble AI-generated hate speech can cause on the internet.  However, Tay was actually intended to interact with all of society and was released intentionally; thus, Tay itself might be better viewed as an instance of Type~3 problems or perhaps Type~5.  Safety issues arising more recently with Microsoft Bing Chat may be viewed similarly.}
\end{story}
 
Obviously, researchers will exercise some level of caution to prevent AI systems and their products from ``getting out'' unexpectedly; otherwise Chernobyl-like disasters can result from failures to contain extremely impactful systems and data.  
 But there are more subtle ways in which an AI technology could end up with a larger scale of impact than its creators anticipated.  For instance:
 
\begin{story}{advicebot}{Indulging Advice Bot}{Story 2b: The Indulging Advice Bot.} A chat-bot is created to help users talk about stressors in their personal life.  A 6-month beta test shows that users claim a large benefit from talking to the bot, and almost never regret using it, so an open source version of the bot is made available online, which can be downloaded and used for free even without an internet connection.  The software ``goes viral'', attracting many more users than expected, until over 50\% of young adults aged 20 to 30 become regular users of the bot's advice. When the bot gives the same advice to multiple members of the same friend group, they end up taking it much more seriously than in the beta tests (which didn't recruit whole groups of friends). As a result of the bot's frequent advice to ``get some distance from their stressors'', many people begin to consider dropping out of college or quitting their jobs.  Ordinarily this would be a passing thought, but finding that many of their friends were contemplating the same decisions (due to the influence of the bot), they feel more socially comfortable making the change.  Many groups of friends collectively decide to leave their jobs or schools.  Public education suffers, and unemployment rates increase.
\end{story}

\paragraph{Analysis.}  In each of the above stories, a technology turns out to have a much larger impact on society that expected, and that impact turns out to be bad.  In the first story, the release of the technology is an accident, whereas in the second story the release is intentional but the manner and scope of adoption was unexpected.
 
Professional standards and ethics have a major role to play in encouraging AI developers to predict and avoid outcomes like this.  There is also technical work to be done: AI systems should be developed with some ability to predict whether actions will be ``high impact'' or ``low impact'' \citep{armstrong2017low}, and to avoid having a greater impact than intended, especially on variables outside the domain of the system's training and expertise.  A variety of impact control concepts have been considered using various definitions of impact, such as by \citet{taylor2016alignment, amodei2016concrete, krakovna2018measuring, huang2019learning, shah2019preferences, turner2020conservative}.  These are preliminary and haven’t been tried much in real-world settings, and in particular have not been applied to natural language systems.

\paragraph{Impact restrictions.} We may wish to treat one or more protected features of society as \emph{outside the domain} of an AI system's allowable influence.  For instance, in the \storyref{advicebot} story above, the significant increase in unemployment rates was very different from and more significant than the kind of impact its creators expected. One way to restrict the impact of an AI system might be to have the system predict and avoid disallow significant impacts outside of its allowed domain.

An AI system could predict and control its own impact on the world in either a ``model-based'' for ``model-free'' fashion.  Prediction and control of a quantity is said to be model-based if it is based on a representation of the world, internal to the system.   Thus, \emph{model-based impact control} could use one of the above definitions of impact directly to predict the ``impact level'' of various actions before making a choice. By contrast, \emph{model-free} control of a quantity is learned from past experiences of what affected the quantity, often in settings with the AI system's designers do not know how to model the system's environment.  \emph{Model-free impact control} could be implemented if an impact metric is calculated by a trusted source external to the AI system, and provided as a signal for the system to observe, predict, and control.  Such solutions might resemble social relationships where people or institutions define boundaries for other agents to respect, without having to explain treasons to those agents (e.g., ``mind your own business'', ``get out of my backyard'').

Human professionals with a heightened capability to influence others---such as doctors, lawyers, and therapists---typically undergo significant training and enculturation to learn what is or is not appropriate for them to influence, and this understanding often depends on at least an amateur knowledge of how the world works outside their field. As such, it may be a very challenging learning problem for advanced AI systems to reliably limit their own impact.
 
\paragraph{Scope sensitivity.} Ideal behavior for an AI system is a function of how many copies of the system have been implemented, and where. For instance, if a bot convinces one person to go to Central Park for a lunch break, a relaxing walk results; but if a million copies of the bot convince a million people to go there all at once, the result is a terribly crowded park.  So, a new AI technology needs to be designed to act differently---and typically more conservatively---based on the number of the instances of the technology that are running, and their context.  In other words, new AI technologies need to be sensitive to the scale on which they are being applied.  This at least requires each implementation to know roughly how many other implementations are out there, which requires at least a minimal degree of communication between implementations, perhaps mediated by human overseers responsible for limiting the scope of a beta test. Without scope sensitivity, the impact of a new AI technology could be much larger than expected, simply as a result of an unexpected degree of popularity.


\subsection{Type~3: ``Worse than expected'' AI impacts}
Oftentimes, the whole point of producing a new AI technology is to produce a large (usually positive) impact on society.  Therefore, a major category of societal-scale risk arises from large, well-intentioned interventions that go wrong.  

The following story illustrates how a messaging assistant technology could learn to cause its users to distrust each other, while the company that creates it has no intention to create that effect:
 
\begin{story}{email}{Cynical Email Helper}{Story 3a: The Cynical Email Helper.}  A tech giant with over a billion users releases a new ``email helper'' feature that reads a user's email and suggests full email responses for the user to send, sometimes multiple paragraphs in length. However, many users struggle to understand the helper’s reasoning behind its messages, so a new feature is added that privately explains to the user why the message might be a good idea.  A typical display to the user looks like this:
\begin{itemize}
\item[] \underline{Message from Julia:} ``Hey, want to come to my party at 8:00 tomorrow?''
\item[] \underline{Suggested response:} ``Sure, Julia, I'd love to come to your event! Is it alright if I arrive a bit late, at 9:00?''
\item[] \underline{Reason for response:} Remember you have plans to meet with Kevin from 5:30 to 8:30, although there's no need to mention that detail to Julia; she might be jealous or offended.
\end{itemize}

 
The helper is programmed to improve over time, from positive feedback whenever the user chooses to send the suggested message.  Ironically, the helper gets more positive feedback when it makes the user more nervous about the situation, such as by pointing out ways the counterparty could get angry at the user.  This pattern causes users to feel like their helper is supporting them through a (purportedly) tricky social situation.  So, the helper learns to gradually include more and more advice that causes users to keep secrets and fear offending each other. As a result, a large fraction of the population becomes gradually more anxious about communicating with others in writing, while also becoming increasingly easy to offend as forthright communication styles become rare. It takes years for everyone to notice the pattern, but by that time many people have become excessively distrustful of others. The creators of the technology wish they had included a user experience question like ``how are you feeling about your email today?'', to measure how their product might be affecting people separately from measuring how much people use it.
\end{story}

In the above story, the tech company did not design the harmful behavior; it was learned.  Such failure modes are not limited to producing psychological harm; consider the following variant of the same story, where the harm is institutional rather than psychological: 

\begin{story}{mediator}{Corrupt Mediator}{Story 3b: The Corrupt Mediator.} A new company that calls itself Mediation.AI\footnote{All company names in the stories of this article are purely fictional. If any such names have also been used by real companies, it is entirely by coincidence, not reference.} releases natural language tools for helping mediate conflicts between large institutions that have overwhelming amounts of communication to manage during negotiations.  Many governments of neighboring jurisdictions and states begin using the software to negotiate laws and treaties.  Like in the previous story, the tool is programmed to learn strategies that increase user engagement, as a proxy for good performance.  Unfortunately, this leads to the software perpetually resolving short-term disputes that relieve and satisfy individual staff members involved in those disputes, while gradually creating ever more complex negotiated agreements between their governments, rendering those governments increasingly dependent on the software to handle foreign affairs.  International trade relations begin a long and gradual decline, which no one country is able to negotiate its way out of.  Frequencies of wars gradually also increase due to diminished incentives to cooperate.
\end{story}

\paragraph{Analysis.}  The previous two stories illustrate how using a technology frequently is not the same as benefiting from it.  To begin paying more direct attention to benefit, let us consider the relationship between one or more human stakeholders and one or more AI systems to whom the humans are delegating tasks or responsibilities, and whether the humans benefit from that relationship.

\paragraph{Single/single delegation.}  The problem of ensuring that a single AI system will benefit (i.e., serve the interests of) a single user is called ``user/agent value alignment'' \cite{shapiro2002user}, or more recently, ``AI alignment'' \citep{soares2014aligning,taylor2016alignment}.  Single/single delegation problems raise numerous subtle ``alignment'' issues, such as:
\begin{itemize}
\item \textbf{deception:} if the system's learning objective is defined entirely by user feedback, it might achieve that objective partly by tricking the user into thinking it's more helpful than it is;
\item \textbf{racketeering:} if the system's learning objective increases with user engagement, it might learn to achieve that objective partly by \emph{racketeering}, i.e., creating novel problems for the user that increase the user's reliance on the system (e.g., debilitating the user, or raising others' expectations of the user).
\item \textbf{self-preservation:} in particular, the system has an incentive to prevent the user from turning it off, which it might achieve by deception or racketeering.
\end{itemize}

Indeed, reinforcement learning systems can in principle learn to manipulate the human minds and institutions in fairly arbitrary (and hence destructive) ways in pursuit of their goals \citep[Chapter 4]{russell2019human} \citep{krueger2019misleading} \citep{shapiro2011social}.  Regulations against false advertising and racketeering laws are important historical examples of how principles of free speech have sometimes been balanced against the negative externalities of widespread deception and manipulation.  Sometimes, user privacy can help protect them from certain forms of manipulation. However, even model-free learning techniques can control hidden state variables in their environments, as demonstrated by any reinforcement learning algorithm for solving unknown POMDPs.

It is possible to somewhat mitigate these issues with reinforcement learning by designing the AI system to solve an \emph{assistance game} with the human (sometimes previously known as a CIRL game) \citep{hadfield2016cooperative}.  An assistance game is a two-player game between the human and the AI system.  The system's objective is to serve the human's preferences, but the system is uncertain about those preferences, and it learns about them over time from the human's behavior.  This problem framing helps to some degree with avoiding deception, racketeering, and self-preservation.  For instance, deceiving the user distorts the system's own access to information about its subjective, which is suboptimal from the system's perspective. Racketeering and self-preservation at the user's expense are similarly poor strategies within the assistance game framework.

However, malfunctions can still occur if the parameters of the assistance game are misspecified \citep{carey2018incorrigibility, milli2019literal}.  Moreover, assistance games in their simplest form do not address the issue that the user's preferences themselves could be changed by the technology \citep[Chapter 9]{russell2019human}.  While some users might endorse their core values being changed by an AI system, others might find the idea horrific.  Appropriately restricting the impact of AI technologies on the human mind poses a significant challenge, particularly because AI technologies are often used primarily to provide information for human consumption.
 
If all that wasn't complicated enough, protecting society as a whole from large-scale intervention malfunctions is a much more complex game than serving a single human, as Stories 3a and 3b above both serve somewhat to illustrate.
 
\paragraph{Multi/single delegation.}  Any plan for ensuring an AI system will benefit society will need to account for the fact that the system's user(s) and creator(s) will simultaneously aim to derive particular benefits from its existence.  This suggests a game with at least four players: the system itself, its creator(s), its user(s), and some representation of the rest of society as one or more players.  Moreover, some AI systems might be explicitly designed to serve many stakeholders at once, such as an office assistant system, or a system designed to aid in public policy decisions.  We call this situation \emph{multi/single delegation}: multiple human stakeholders depending on a single AI system to fulfill a purpose.
 
\paragraph{Multi/multi delegation.}  There is always the possibility that many separate optimization processes (either AI systems, or human-AI teams) can end up in a Prisoner's Dilemma with each other, each undoing the others' efforts by pursuing its own.  Thus, in the end we will need a good formalism in which many stakeholders can be served simultaneously by many AI systems, i.e., \emph{multi/multi delegation}.  Such a formalism would no doubt aid in addressing the other problems raised in this article as well.

\subsection{Type~4: Willful indifference}
All of the potential harms in the previous sections are made more likely if the creators of AI technology are unconcerned about its moral consequences.  Even if some employees of the company detect a risk of impacts that's bigger than expected (Type~2) or worse than expected (Type~3), it may be quite difficult to institute a change if the company is already profiting greatly from its current strategy, unless there is some chance of exposure or intervention from outside the company to motivate a reform.  The following story illustrates this:

\begin{story}{abtesting}{Harmful A/B Testing}{Story 4: Harmful A/B Testing.} A tech company called X-corp uses an automated ``A/B testing'' system that tries out new parameter values to expand its user base.  
Like in the \hyperref[story:mediator]{Corrupt Mediator} 
story, their system learns that they can get more users by causing their users to create problems for each other that only X-corp's tools can solve, creating a powerful network effect that rapidly expands X-corp's user base and earns X-corp a lot of money.  
Some concerned X-corp employees complain that they have inadequate checks in place to ensure their A/B development process is actually benefiting their users, but it never seems to be a convenient time to make major changes to the company's already profitable strategy.  
One employee manages to instigate an audit from a external non-profit entity to assess the ethics of X-corp's use of AI technology.  
However, X-corp's A/B testing system is opaque and difficult to analyze, so no conclusive evidence of ethical infractions within the company can be identified.  
No regulations exist requiring X-corp's A/B testing to be intelligible under an audit, and opponents of the audit argue that no technology currently exists that could make their highly complex A/B testing system intelligible to a human.  
No fault is found, and X-corp continues expanding and harming its user base.
\end{story} 

\paragraph{Analysis.} This story spells out how our collective strategy for preventing societal harm must go beyond merely providing methods that allow the building of safe and beneficial AI technology.  We must also establish these methods as worldwide industry standards and norms that cannot be ignored.  Industry norms are usually maintained by professional codes of conduct, regulatory bodies, political pressures, and laws.  For instance, technology companies with large numbers of users could be expected to maintain accounts of how they are affecting their users' well-being.  This is primarily not a technological challenge, but a rather, a challenge of establishing a new social contract where, like food and drug companies, companies who deploy interactive algorithms must be continually examined for their impact upon people and society.  Academically, this is a matter for social scientists who study the impact of technology. 

However, there are also opportunities for AI to assist humans in regulating AI technology \citep{raji2020closing}.  Ensuring AI systems make decisions in a manner that is interpretable to humans will be key to this objective, and will limit opportunities for morally indifferent creators to ``look the other way'' when their systems are liable to cause societal-scale harm.

\paragraph{Interpretability techniques.}  A successful audit of a company's business activities requires the company's personnel to understand those activities.  When those activities are automated with AI technology, the actions of the AI systems must themselves be interpretable by company personnel and explainable to outsiders.  ``Black-box'' machine learning techniques, such as end-to-end training of the learning systems, are so named because they produce AI systems whose operating principles are difficult or impossible for a human to decipher and understand in any reasonable amount of time.  Hence, alternatives or refinements to deep learning are needed which yield systems with comparable performance while being understandable to humans.  This requires attention to the amount of information that can be consumed and interpreted by a human \citep{olah2018building}.   \citet{rudin2019stop} argues further that ``trying to explain black box models, rather than creating models that are interpretable in the first place, is likely to perpetuate bad practices and can potentially cause catastrophic harm to society''.  Subsequently, \citet{semenova2019study} provides a technical argument that very little performance may need to be sacrificed to drastically improve interpretability.  Work in this direction could be very useful to maintaining accountability for companies engaged in highly automated business activities.

\subsection{Type~5: Criminal weaponization}
It's not difficult to envision AI technology causing harm if it falls into the hands of people looking to cause trouble, so no stories will be provided in this section.  It is enough to imagine an algorithm designed to pilot delivery drones that could be re-purposed to carry explosive charges, or an algorithm designed to deliver therapy that could have its goal altered to deliver psychological trauma.  What techniques exist for preventing AI systems from being intentionally modified for harmful purposes?

As an industry-ready example, suppose AI researchers have developed a scene description tool $D:scenes \to paragraphs$, which takes as input an image of a potentially complex scene, and returns a paragraph of text that accurately describes what is happening in the scene. Now suppose we want to release the tool for public use.  However, to prevent it from being used freely to target or study individuals, we wish to block public users of $D$ from using it to describe certain types of scenes, such as a scene containing a person, or a scene that has been digitally altered (such as to add or remove a person).  A naive approach might be to train a new version, $D'$, on data that contains no unacceptable scenes, and hope that the trained algorithm would perform poorly on queries to describe unacceptable scenes.  But, this hope might not pan out if the learned function turns out to generalize well to unacceptable examples.  And, if the training process is very computationally expensive, it won't be easy to repeat.

A better approach would be to use \emph{program obfuscation}.  Before releasing $D$, we could train another (simpler) function $A: scenes \to \{true,false\}$ for detecting whether a scene image is acceptable for the software to describe.  We'd then write a new function, $SD = \texttt{safe\_descriptor}: images \to labels$, like this: 
\begin{verbatim}
import D as description, A as acceptability
SD = safe_descriptor = function(scene):
    if acceptability(scene) = true:
        return description(scene)
    else:
        return ``unacceptable scene''
\end{verbatim}

Of course, it may be relatively easy for a hacker to ``take apart'' a compiled version of SD, and run the \texttt{description} subroutine without the \texttt{acceptability} check.  This is why we need to ``obfuscate'' SD.  An ``obfuscation'' function $Ob: programs \to programs$ returns a new program $Ob(SD)$ to be released instead.  The (compiled) code of $Ob(SD)$ is mangled and so that it cannot be easily ``taken apart'', but it computes the same input/output function as $SD$.

Historically, there have been many ad hoc obfuscation methods employed by software companies to protect their intellectual property, but such methods have a history of eventually being broken \citep{barak2002can}.
To prepare for a future with potentially very powerful AI systems, we need more rigorously proven methods.  Luckily, there has been recent progress in cryptography developing theoretical foundations for a technique called \emph{indistinguishability obfuscation (IO)} \citep{garg2016candidate,lin2016iofromddh,bitansky2018indistinguishability}, which can be used to implement $Ob$ for the purpose above \citep{garg2016hiding}.  While these methods are currently too inefficient to be practical, this area of work seems promising in its potential for improvements in speed and security.  This leaves open a rich domain of problems relevant to AI and cryptography:

\begin{enumerate}
\item Can IO techniques be made more efficient for obfuscating a specific class of AI-relevant programs, such as neural networks or bounded-depth probabilistic programs?
\item Can new or existing IO techniques be shown to work under more secure cryptographic assumptions?  While a purely cryptographic question, a positive answer to this would increase our credence that IO techniques will not be broken by AI systems in the future.
\end{enumerate}
\subsection{Type~6: State weaponization}  
Tools and techniques addressing the previous section (weaponization by criminals) could also be used to prevent weaponization of AI technologies by states that do not have strong AI research labs of their own.  But what about more capable states?

The elephant in room here is that AI can be used in war.  Some argue that, ideally, mechanical drones could be pitted against one another in casualty-free battles that allow nations to determine who would win a war of lethal force, without having to actually kill any human beings.  If taken no further, this would be a major improvement over current warfare practices.  However, these capabilities are not technologically far from allowing the mass-killing of human beings by weaponized drones.  Escalation of such conflicts could lead to unprecedented violence and death, as well as widespread fear and oppression among populations that have been targeted by mass killings.

It may seem that the only action computer scientists can take to prevent such outcomes is to refuse participation in the design of lethal autonomous weapons.  Is there anything positive we can contribute to the age-old problem of world peace?

Although it may be a long shot, it's conceivable that AI technology could be employed to \emph{eliminate or reduce incentives} for states to engage in war.  For instance, AI could make it easier to share resources, by brokering mutually agreeable peace treaties.  Or, technical solutions for sharing control of powerful AI systems could help to prevent wars from emerging over how those AI systems should be used.  While any given attempt to use AI technology to resolve global conflicts is unlikely to succeed, the potentially massive upside makes this possibility worth exploring.  For instance, there are currently numerous open technical problems in how to approach AI-assisted negotiation, and the examples below are far from exhaustive.

\paragraph{Mediation tools.} Consider two countries that would benefit from a peace treaty or trade agreement, but are struggling to reach agreement on the terms.  Or, imagine two friends who can't agree on where to have dinner.  As a prerequisite for an AI system to propose a compromise solution in such a scenario, we need AI technology capable of formulating a plan that one party finds acceptable and the other can understand.  

For PhD-level work in this area, consider the following cooperative online game between Alice (human), Bob (human), and an AI assistant Medi.  Bob has access to a video game screen and controller, but the goal of the game hidden from him.  
Alice is on the other side of the world, and can see the goal, but doesn't have access to the controller.  Alice is allowed to convey messages to Bob about the video game goal; she can write her own message and pay an in-game cost (representative of the cost of writing an email), or choose from a list of suggested messages written by Medi (at no cost).  At first Alice's own written messages to Bob will be much better than Medi's, but with a lot of practice on various (Alice,Bob,videogame) scenarios, can we train Medi to start providing valuable low-cost suggestions to Alice?

Formally, we can view Alice, Bob, and Medi as solving an instance of a \emph{Decentralized POMDP} \citep{bernstein2002complexity}, $\langle S,A_1,A_2,A_3,P,R,\Omega_1,\Omega_2,\Omega_3, O,T,K\rangle$, where~$A_1$ is Alice's action space (choosing a message from the assistant's presented options, or writing her own and paying the cost), $A_2$ is Bob's action space (moving the game sprite), and $A_3$ is Medi's action space (displaying lists of message options for Alice to choose from).  The team's score in the game, $R$, is defined by Bob's score in the single-player video game minus the attentional cost of the messages Alice wrote.  So, Medi does a good job if she conveys useful information from Alice to Bob, at low attentional cost to Alice.

If we can develop good solutions to this sort of problem, numerous possibilities open up, including potentially saving Alice a lot of time on writing emails.  But to push the science specifically toward better mediation tools, a natural next step would be to try experiments with with a symmetrized version of the game, where both Alice and Bob have goals and can take actions that affect both of their goals, and are assisted by an AI mediator Medi who can write suggested messages for both of them.  Medi could sometimes send a message to Alice and Bob simultaneously, to create a ``contract'' between them if they both agree to it.  

\paragraph{Negotiable controls for powerful systems.}
In order to reduce the risk of conflict over the control of powerful AI systems or other systems, it would be prudent to develop formal, AI-compatible principles for sharing control of powerful processes.

There is an interesting tension in this area, between fairness and successful negotiation.  Suppose Alice and Bob are negotiating a deal to control a powerful system, and a mediator Medi is assisting in the negotiation.  
Medi may be able to finalize the deal by proposing a plan that's great for Alice but potentially terrible for Bob, in a way that Bob is unable to recognize in advance.  (Betting is a simple example of this: a bet looks good to both parties, but can only carry positive expected value for one of them in reality.)  This seems somewhat unfair to Bob.  On the other hand, if Medi doesn't propose plans that look appealing from Bob's subjective perspective, Bob might walk away from the bargaining table.  

Hence, there is sometimes a fundamental trade-off between a deal looking good to both Alice and Bob, and the deal treating Alice and Bob equitably over time \citep{critch2017servant}.  This trade-off can be seen in the behavior of reinforcement learning systems that are Pareto optimal for principals with different beliefs 
\citep{critch2017toward,desai2018negotiable}.  The only way to eliminate this trade-off is to eliminate the differences in beliefs between the principals. For that, perhaps progress in building mediation tools would be a useful start, or control techniques for powerful AI systems that can explicitly account for differences in beliefs among a committee of humans controlling a single system, such as in Dalrymple's ``Open Agency Architecture'' concept \citep{dalrymple2022open}.
 
\section{Conclusion}

At this point, it is clear that AI technology can pose large-scale risks to humanity, including acute harms to individuals, large-scale harms to society, and even human extinction.  Problematically, there may be no single accountable party or institution that primarily qualifies as blameworthy for such harms (Type~1).  Even when there is a single accountable institution, there are several types of misunderstandings and intentions that could lead it to harmful outcomes (Types 2-6).  These risk types include AI impacts that are bigger than expected, worse than expected, willfully accepted side effects of other goals, or intentional weaponization by criminals or states.  For all of these risks, a combination of technical, social, and legal solutions are needed to achieve public safety.

\cleardoublepage
\bibliography{main}
\end{document}